\documentclass[10pt,twocolumn,letterpaper]{article}
\usepackage[pagenumbers]{cvpr}
\definecolor{MyDarkRed}{rgb}{0.8,0.02,0.02}
\definecolor{MyDarkBlue}{rgb}{0.02,0.02,0.8}
\definecolor{MyDarkGreen}{rgb}{0.1,0.8,0.1}

\newcommand{\model}{Presto\xspace}
\newcommand{\dataset}{LongTake-HD\xspace}
\newcommand{\sca}{Segmented Cross-Attention\xspace}

\definecolor{cvprblue}{rgb}{0.21,0.49,0.74}
\usepackage[pagebackref,breaklinks,colorlinks,allcolors=cvprblue]{hyperref}
  
\usepackage[accsupp]{axessibility}
\usepackage{multicol}
\usepackage{multirow}
\usepackage{float}
\usepackage{listings}
\usepackage{array}
\usepackage{calc}
\usepackage{cuted}
\newcolumntype{P}[1]{>{\centering\arraybackslash}p{#1}}
\usepackage{amssymb}
\usepackage{pifont}

\usepackage{appendix} % Important: Include the appendix package

\definecolor{codegreen}{rgb}{0,0.6,0}
\definecolor{codegray}{rgb}{0.8,0.8,0.8}
\definecolor{codepurple}{rgb}{0.58,0,0.82}
\definecolor{backcolour}{rgb}{0.97,0.99,0.97}
\lstdefinestyle{mystyle}{
    backgroundcolor=\color{backcolour},   
    commentstyle=\color{codegreen},
    keywordstyle=\color{magenta},
    numberstyle=\tiny\color{codegray},
    stringstyle=\color{codepurple},
    basicstyle=\ttfamily\scriptsize,
    breakatwhitespace=false,         
    breaklines=true,      
    breakautoindent=false,
    breakindent=0ex,
    captionpos=b,                    
    keepspaces=true,                 
    numbers=left,                    
    numbersep=5pt,                  
    showspaces=false,                
    showstringspaces=false,
    showtabs=false,                  
    tabsize=2,
    showlines=true
}
\def\paperID{2999} % *** Enter the Paper ID here
\def\confName{CVPR}
\def\confYear{2025}

\title{Long Video Diffusion Generation with Segmented Cross-Attention and Content-Rich Video Data Curation}

\author{Xin Yan\thanks{Work done while interning at 01.AI. Contact: cakeyanxin@gmail.com}\quad Yuxuan Cai\quad Qiuyue Wang\quad Yuan Zhou\quad Wenhao Huang\quad Huan Yang\thanks{Correspondence to: hyang@fastmail.com} \\
01.AI\\
}

\begin{document}

\maketitle
\begin{strip}
\begin{minipage}{\textwidth}\centering
\vspace{-10pt}
\includegraphics[width=\linewidth]{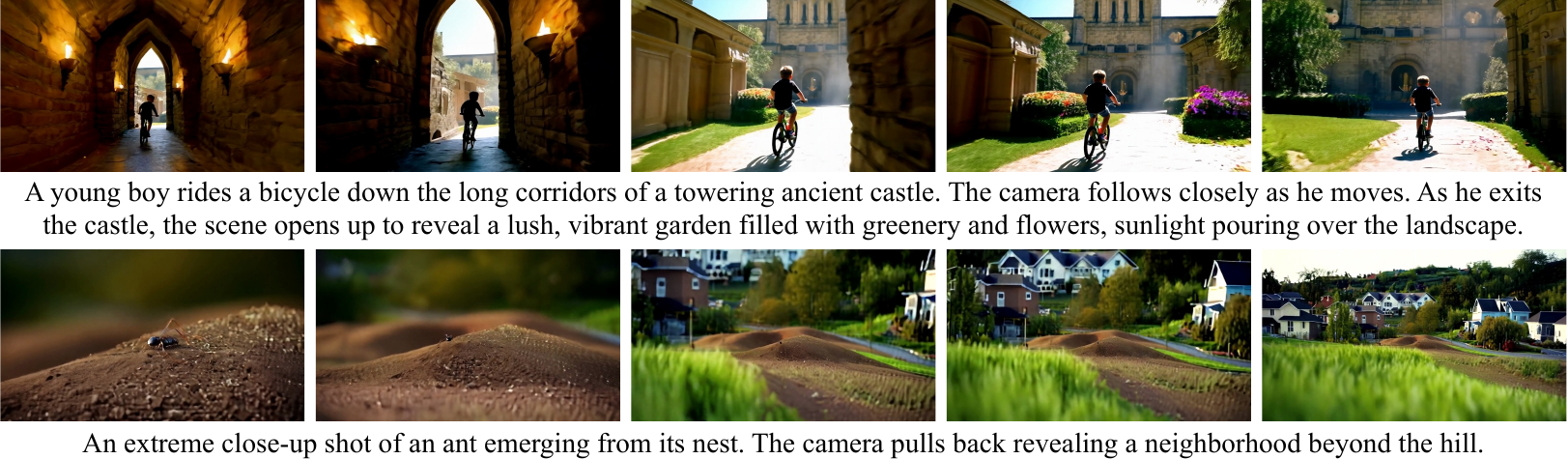}
    \captionof{figure}{\model can generate long videos with rich content and long-range coherence.}
    \label{fig:teaser}
\end{minipage}
\vspace{10pt}
\end{strip}

\begin{abstract}
We introduce \textbf{\model}, a novel video diffusion model designed to generate 15-second videos with long-range coherence and rich content. 
Extending video generation methods to maintain scenario diversity over long durations presents significant challenges.
To address this, we propose a \sca (SCA) strategy, which splits hidden states into segments along the temporal dimension, allowing each segment to cross-attend to a corresponding sub-caption.
SCA requires no additional parameters, enabling seamless incorporation into current DiT-based architectures.
To facilitate high-quality long video generation, we build the \dataset dataset, consisting of 261k content-rich videos with scenario coherence, annotated with an overall video caption and five progressive sub-captions. 
Experiments show that our \model achieves 78.5\% on the VBench Semantic Score and 100\% on the Dynamic Degree, outperforming existing state-of-the-art video generation methods.
This demonstrates that our proposed \model significantly enhances content richness, maintains long-range coherence, and captures intricate textual details.  More details are displayed on our project page: \href{https://presto-video.github.io/}{presto-video.github.io}.

\end{abstract}
    
\section{Introduction}
\label{sec:intro}

Video diffusion models~\cite{ho2022video, singer2022make, voleti2022mcvd, ruan2023mm, wang2024videocomposer} have shown an impressive ability to generate high-quality videos based on a single text prompt~\cite{saharia2022photorealistic, nichol2021glide}. However, most current approaches primarily focus on generating short video clips ranging from 3 to 8 seconds, limiting the expressiveness and richness of the resulting content.

To generate longer videos, early approaches incorporate an additional interpolation or extrapolation phase to extend short clips, using techniques like noise scheduling~\cite{wang2023gen, qiu2023freenoise, kim2024fifo} or attention manipulation~\cite{lu2024freelong, tan2024video}. While these methods work well for generating minute-long videos, they struggle to extend beyond the scene content, as they are constrained by the limited capacity of the original short clips.
An alternative approach adopts a more direct strategy, typically by adding new modules to extend the video length in an auto-regressive manner~\cite{yin2023nuwa, henschel2024streamingt2v, wang2024loong}. However, this introduces the challenge of error propagation.

Unlike short clips, long videos require a balance between content diversity and long-range coherence, posing significant challenges for current video generation methods.
To generate long videos with rich content, we recognize the importance of expanding the text input by incorporating multiple texts, as demonstrated in previous approaches~\cite{oh2023mtvg, bansal2024talc}. While combining videos generated from each text can enhance content diversity, it often leads to abrupt transitions between different scenarios.
One alternative is to incorporate multiple texts into the video generation model simultaneously.
This method provides the model with a broader range of textual inputs, enabling the generation of more content-rich videos, in contrast to the traditional approach of using a single text input, which limits the available information.
This approach helps ensure long-term coherence in the generated videos by modeling the texts concurrently, providing a seamless viewing experience.

Existing long video generation methods overlook the importance of high-quality data~\cite{bansal2024talc,henschel2024streamingt2v}, leading to low consistency and content diversity in generated videos.
A large-scale, high-quality video dataset is crucial to obtaining a model that generates content-rich and coherent long videos. This dataset should include long videos with long-range coherence and complex dynamics, along with multiple distinct yet coherent textual descriptions for each video.
To this end, we develop a systematic data curation pipeline to collect content-rich video-prompt pairs from public datasets, contributing to creating our \textbf{\dataset} dataset. 
We filter 261k single-scene video clips from 8.9M publicly accessible videos as the pre-training set.
Then, we apply additional meticulous filtering steps to select the finest instances, resulting in a fine-tuning dataset of 47k clips. This refined dataset ensures that our method can generate high-quality, extended-duration, and content-rich videos.

To tackle the challenge of long videos, we propose \textbf{\model}, a novel method capable of generating 15-second videos with \textbf{rich content} and \textbf{long-range coherence}, as shown in \cref{fig:teaser}.
Instead of using a single long caption for each video, we divide the visual content into segments and generate progressive sub-captions that align with the unfolding storyline.
Next, we modify the cross-attention mechanism in the Diffusion Transformer (DiT).
To adapt the DiT model for long video generation, we refine the text embedding process and the cross-attention mechanism to effectively handle multiple progressive text conditions alongside temporal information. In particular, we introduce \textbf{\sca (SCA)}, which divides the hidden states into segments along the temporal dimension and cross-attends each segment with its corresponding sub-caption. SCA introduces no additional parameters or modules and can seamlessly integrate into existing DiT-based methods with minimal fine-tuning.
We explore three distinct SCA strategies to manage the interaction between text embeddings and segmented latent features: Isolate \sca (ISCA), Sequential \sca (SSCA), and Overlap \sca (OSCA). Our experiments demonstrate that OSCA enhances the content richness and long-range coherence in the generated long videos.

Experimental results demonstrate the effectiveness of our methods. \model achieves a 78.5\% score on the VBench Semantic Score, outperforming both the leading open-source model, Allegro, and the commercial system, Gen-3.
Notably, \model achieves a perfect 100\% score on the Dynamic Degree metrics, showcasing its outstanding ability to capture dynamics and transitions.
The quantitative results highlight our approach's strength in capturing intricate textual details and generating videos with rich content.
Furthermore, our user study indicates that \model excels in scenario diversity, scenario coherence, and text-video alignment when compared to various open-source and commercial-level works.

Our key contributions are outlined as follows: 
\begin{itemize}
    \item We propose a large-scale video dataset \textbf{\dataset}, with 261k high-quality cases curated from publicly sourced videos, characterized by long-duration, content-rich, and long-range coherent videos, and each paired with progressive sub-captions. 
    \item We proposed \textbf{\model} for long video generation with a simple yet effective \textbf{\sca} strategy which extends the DiT architecture to accommodate multiple text prompts, enabling the generation of videos with rich content and long-range coherence. 
\end{itemize}

\section{Related Work}\label{sec:related}

\noindent\textbf{Long-Video Generation} is a challenging task in video generation, requiring videos to be both rich and coherent.
One paradigm~\cite{lu2024freelong,tan2024video,wang2023gen,qiu2023freenoise,kim2024fifo} is to modify noise scheduling or attention mechanisms during the inference stage, which are highly constrained by the original video clips, often resulting in limited content diversity.
Another approach~\cite{yin2023nuwa,henschel2024streamingt2v,wang2024loong} directly generates long videos, by introducing additional modules with auto-regressive generation, requiring substantial training resources.

\noindent\textbf{Multiple-Text-to-Video Generation (MT2V)}~\cite{bansal2024talc,oh2023mtvg,ramos2024contrastive, villegas2022phenaki} aligns with our work, as we also incorporate multiple text prompts. 
While MT2V aims to create videos from multiple inputs, our model uniquely adheres to the traditional T2V framework by requiring only a single user prompt. 
Additionally, MT2V methods such as TALC~\cite{bansal2024talc}, typically combine multi-scene captions rigidly (\eg, “A panda is running in the park, sunny.” and “A golden retriever is running in the park, autumn.”). Our progressive caption method eliminates redundant descriptions across sub-captions and focuses on scenario transitions. That means our multiple sub-captions can coalesce into a continuous narrative, improving the coherence of the video content. 

\noindent\textbf{Video Generation Datasets} are crucial for pre-training high-quality video generation models. 
Existing text-video datasets~\cite{bain2021frozen} have made substantial progress in terms of dataset size, such as Panda-70M~\citep{chen2024panda}, HD-VILA~\citep{xue2022advancing}, and HD-VG~\citep{wang2023videofactory}, which contain 70M, 100M, and 130M video clips respectively. Recent works such as OpenVid-1M~\citep{nan2024openvid} and FlintstonesHD~\cite{yin2023nuwa} have attempted to construct small, yet higher-quality datasets. 
In contrast, we propose \dataset, focusing on the finest quality videos with rich content, long-range scenario coherence, and multiple progressive sub-captions per video.

\noindent\textbf{Time-Varying Text Prompts.} Notably, works like Phenaki~\cite{villegas2022phenaki} and VideoPoet~\cite{kondratyuk2023videopoet} also explore the idea of utilizing the time-varying text prompts or latent to generate long videos, aligning with our methodology in high-level. A key difference is that our method presents a comprehensive solution to the long video generation problem, encompassing the dataset, model, and interpolation techniques, while these works primarily focus on the model aspect. 
\section{\dataset Dataset}
\label{sec:data}

\begin{table*}
  \centering
    \renewcommand{\arraystretch}{1.1}
  \renewcommand{\aboverulesep}{0pt}
  \renewcommand{\belowrulesep}{0pt}
  \setlength{\tabcolsep}{0pt}
      \resizebox{1\linewidth}{!}{
  \begin{tabular}{p{5.8em}|P{5.5em}P{5.5em}P{5.5em}|P{5em}P{5em}P{5em}P{5em}P{5em}}
    \toprule
    \hspace{4pt} Dimensions & \multicolumn{3}{c|}{Video Captions} & \multicolumn{5}{c}{Videos} \\
      \midrule
    \hspace{4pt} Dataset & Caption & Sub-Captions & Tokens & Duration & Aesthetics$^\dag$ & Diversity$^{\dag\ddag}$ & Coherence$^{\dag\ddag}$ & Quality$^{\dag\ddag}$ \\
      \midrule
   \hspace{4pt} Panda & \ding{51} & \ding{55} & 13.2 & 8.5s & 4.62 & 2.55 & 2.38 & 2.18 \\
   \hspace{4pt} HD-VILA & \ding{51} & \ding{55} & 32.5 & 13.4s  & 4.78 & 2.52 & 2.49 & 2.31 \\
      \midrule
   \hspace{4pt} Ours & \ding{51} & \ding{51} (5) & \bf 186.42 & \bf 15.7s & \bf 5.21 & \bf 3.02 & \bf 3.44 & \bf 2.80 \\
    \bottomrule
  \end{tabular}
  }
  \caption{Comparisons between popular text-video datasets and our LongTake-HD on different dimensions. Unless specifically noted otherwise, data is calculated over the entire dataset using automated metrics. Our dataset leads in all dimensions. $^{\dag}$: These aspects are evaluated on 100 random samples. $^{\ddag}$: These aspects are evaluated via human reviews on a four-point scale. }
  \label{tab:comp}
\end{table*}

Data curation plays a crucial role in training our proposed model. Both high-quality video content and detailed descriptive captions are essential for generating content-rich videos with long-range coherence. 
Beginning with a dataset of 8.9 million publicly available raw videos, we filtered it down to single-scene video clips that exhibit diverse content, with a resolution of 720×1280, a minimum duration of 15 seconds, and high aesthetic quality. 
This process yielded 261k instances, each comprising a content-rich video paired with five coherent and progressively structured sub-captions, as well as an overall video caption. 
To further enable dataset stratification for different training stages, we utilized the full set of 261k instances during the pre-training phase and applied rigorous filtering criteria to extract the finest quality 47k instances for the fine-tuning set.
We refer to this curated dataset as \dataset. 
Further details are shown in \cref{app:details}.

\subsection{Collecting Content-Diverse Video Clips}
\label{sec:data:videos}

Existing publicly available datasets, such as HD-VILA-100M~\cite{xue2022advancing}, Panda 70M~\cite{chen2024panda}, and WebVid-10M~\cite{bain2021frozen}, offer a vast array of diverse and comprehensive video data that reflect the natural distribution of real-world content. However, these raw datasets often contain substantial amounts of noisy and low-quality material, lacking in careful curation for content quality and caption coherence. Inadequate data curation processes can lead to the inclusion of noisy, disjointed, or irrelevant video data, which negatively impacts the training of models, particularly for long-form videos.

Starting from a dataset of 8.9 million publicly accessible raw videos, we apply a video data filtering pipeline including: (1) duration, speed, and resolution filtering; (2) scene segmentation; (3) low-level metrics filtering; and (4) aesthetic and motion contents filtering.

\noindent\textbf{Duration, speed, and resolution.} We exclude videos shorter than 15 seconds, to ensure an adequate video length. To maintain smoothness, we filter out videos with a frame rate lower than 23 FPS. We also remove videos with resolutions below 720p to preserve the visual quality of the dataset. Additionally, videos with aspect ratios less than 1 (\ie, vertical videos) are excluded to ensure consistency throughout the dataset.

\noindent\textbf{Scene segmentation.} We detect the scene cuts in videos and filter out those with abrupt transitions using PySceneDetect~\cite{PysceneDetect}. To further strengthen the process and remove any lingering cuts or transitions, we manually discard the first and last 10 frames of each clip. After completing the scene segmentation, the remaining data filtering is applied to the individual single-scene video clips.

\noindent\textbf{Low-Level metrics.} We use brightness and artifacts as key metrics for low-level filtering. We compute the average grayscale value of video frames and remove those that are overly dark or bright. Detection tools~\cite{watermark-detection,baek2019character} are applied to identify artifacts, such as watermarks and text, that are unrelated to the actual video content.

\noindent\textbf{Aesthetics and motion contents.} We employ the LAION Aesthetics Predictor~\cite{schuhmann2022laion} to evaluate the aesthetic quality of video frames and remove those with low aesthetics scores. Optical flow is calculated using Unimotion~\cite{xu2023unifying}, and clips with higher flow scores are retained to ensure a substantial level of motion amplitude. Furthermore, we compute the coefficient of variation for all optical-flow values within each clip, defined as the ratio of the standard deviation to the mean~\cite{everitt2010cambridge}. This standardized measure of dispersion allows us to assess the smoothness and consistency of motion dynamics, helping to avoid abrupt shifts in motion intensity.

\subsection{Obtaining Coherent Video Captions}
\label{sec:data:caption}

We apply captioning techniques to both images and videos. First, we conduct image-level diversity filtering based on the sampled keyframes. Next, we generate captions for each keyframe and perform semantic filtering on these captions. Finally, we leverage Large Language Models to create multiple progressive sub-captions that include camera motions.

\noindent\textbf{Diversity filtering for keyframes.}
We employ a comprehensive approach to evaluate the diversity and coherence of images, using a combination of low-level and semantic metrics. Specifically, we apply the Peak Signal-to-Noise Ratio (PSNR)~\cite{huynh2008scope} for pixel-wise filtering, the Structural Similarity Index Measure (SSIM)~\cite{wang2004image} for structure-wise filtering, and the Perceptual Similarity (LPIPS)~\cite{zhang2018unreasonable} for semantic-wise filtering.
Additionally, some sampled frames may contain minimal information, such as black screens or blurry images. To address this, we use the image file size as a filtering criterion since frames with low information content typically result in smaller file sizes when compressed in the PNG format~\cite{ziv1977universal, deutsch1996deflate, ziv1978compression, huffman1952method}.

\noindent\textbf{Semantic filtering from captions.}
We utilize Large Vision-Language Models as caption generators to create detailed descriptions for both the entire video and its sampled frames. The captions for individual frames offer in-depth descriptions of the visual elements in each keyframe and represent the corresponding short video segments, while the video caption emphasizes the dynamics and transitions across the video, incorporating both spatial and temporal details.
To generate embeddings for all keyframe captions, we employ MPNet~\cite{song2020mpnet} from SentenceTransformers~\cite{reimers-2020-multilingual-sentence-bert, reimers-2019-sentence-bert} and compute the cosine similarity~\cite{singhal2001modern} between each pair of captions. Additionally, we filter out negative captions, which occur when LVLMs fail to generate responses for frames that contain sensitive or abstract content.

\noindent\textbf{Progressive sub-captions generation strategy.} 
We propose a progressive sub-captions generation approach to create coherent and non-redundant sub-captions that align with the video storyline. We show a simple example of three generated progressive sub-captions below:

{\small
\textit{\textbf{sub-caption 1:} A close-up shot of the ground, focused on a small, slightly elevated, textured mound of soil. A single ant emerges from a tiny opening at the top of the mound, its tiny antennae gently probing the air as it moves cautiously forward.}

\textit{\textbf{sub-caption 2:} The camera gradually pulls back, maintaining focus on the ant as it traverses the surface of the mound.}

\textit{\textbf{sub-caption 3:} Continuing to pull back, the ant diminishes in prominence and focus transitions to reveal a larger view of the immediate ground area.}
}

As in the example above, the first sub-caption describes the main subject, \textit{ant}, and the environment, \textit{soil}. When it comes to the second and third, it continues the previous story and elaborates further on the transitions, \eg \textit{traverses the surface of the mound}, and \textit{a larger view of the immediate ground area}. This narrative-style sub-caption annotation strategy helps enhance the long-range coherence of the generated videos and distinguishes our approach from existing video datasets and MT2V methods. 

For a given long video, we first divide the video clip into $N$ segments and generate independent captions for each segment using our captioning model. We also obtain an overall description of the entire video, capturing both spatial and temporal details.
Next, we refine each sub-caption using a causal approach. We employ an LLM to adjust each sub-caption, considering all previous sub-captions together with the overall description, ensuring that each sub-caption represents a distinct episode within the broader storyline. Additionally, we explicitly incorporate camera motion to enable fine-grained control over the camera. This strategy results in a set of coherent and progressively linked sub-captions for diffusion model training.

During inference, when given a short prompt from the user, we also use the LLM as a ``director" to generate $N$ scripts with consistent and detailed descriptions. For captioning over videos and images, we use Aria~\cite{li2024ariaopenmultimodalnative} as our captioner and GPT-4o~\cite{achiam2023gpt} as the LLM for refining the sub-captions. The detailed prompt templates for these two models are provided in \cref{lst:aria} and \cref{lst:gpt} of the Appendix. We also use an experiment to show the benefits of progressive sub-captions that may enhance the semantics in generated contents in \cref{app:prog} and \cref{tab:textsim}.

\begin{figure*}[htp]
    \begin{center}
    \centerline{\includegraphics[width=1\linewidth]{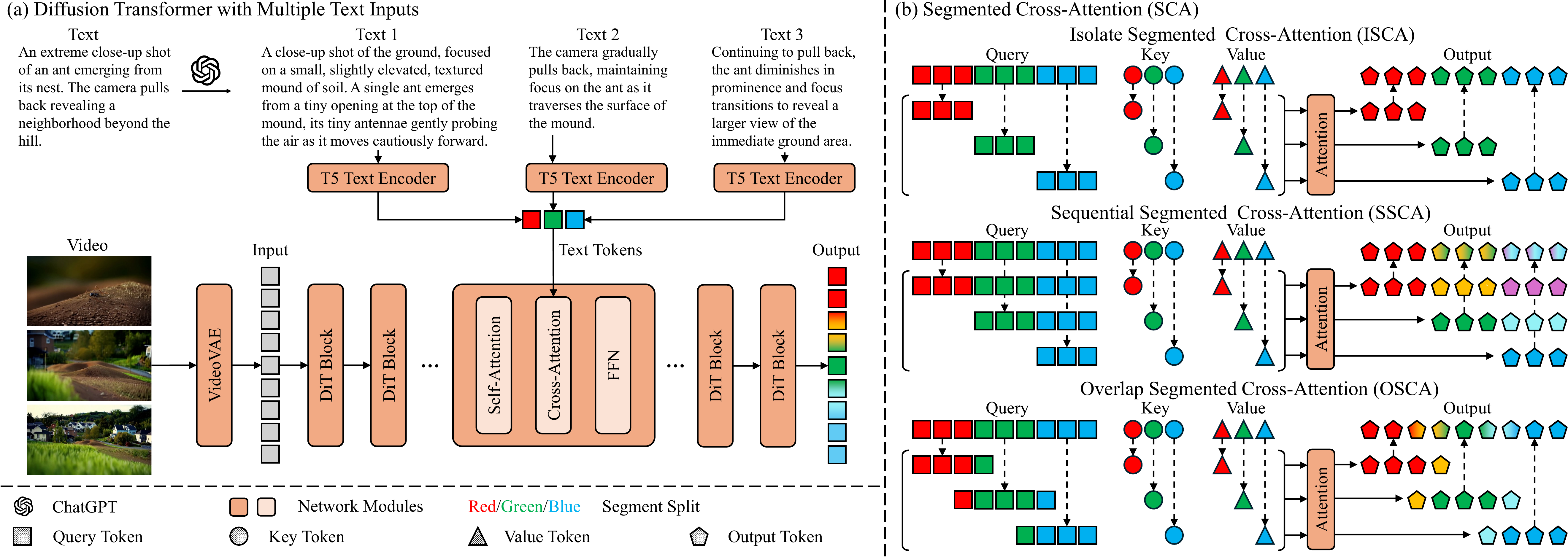}}
    \caption{
    {
    (a) The overall architecture of our \textbf{\model}, which integrates multiple text inputs concurrently. (b) The Segmented Cross-Attention strategy has three variants: 1) Isolated Segmented Cross-Attention (ISCA) directly splits the hidden states along the temporal dimension. The output is concatenated by multiple segments' output. 2) Sequential Segmented Cross-Attention (SSCA) where each segment will see all the previous text conditions. All the overlapped regions are averaged and concatenated with other regions. 3) Overlap Segmented Cross-Attention (OSCA) that is adopted in our method. Only frames at the segment boundary will cross-attend with multiple text conditions.
    } 
    }
    \label{fig:model}
    \vspace{-20pt}
    \end{center}
    \end{figure*}

\subsection{Comparisons between Video Datasets}\label{app:comparisons}

We compare our dataset with some popular text-video datasets, including HD-VILA-100M~\cite{xue2022advancing} and Panda-70M~\cite{chen2024panda}. We evaluate both video captions and videos to show the high quality of our \dataset. Results and details are shown in \cref{tab:comp}.

\noindent\textbf{For video captions}, while existing datasets typically offer a single overall video caption, our dataset includes an additional set of five time-varying sub-captions. These sub-captions are a key distinguishing feature of our \dataset compared to others. Furthermore, these sub-captions can be directly concatenated to form a longer, comprehensive caption, aligning with the format of most text-video datasets. We also calculated the average number of tokens per video. The caption length significantly outperforms other datasets, being about six times longer than HD-VILA and fourteen times longer than Panda. We anticipate this characteristic will particularly benefit Diffusion model training, as highly descriptive captions have been proved crucial for text fidelity and video quality~\cite{videoworldsimulators2024}.

\noindent\textbf{For videos}, we began by calculating the average video duration. Our dataset demonstrates a significant advantage in average video length, primarily attributed to the exclusion of videos shorter than 15 seconds. Furthermore, we conducted a detailed analysis of video quality. Recognizing the complexity of video quality assessment, we concentrated our investigation on four key aspects: Aesthetics, Diversity, Coherence, and Quality. We randomly sampled 100 videos from each dataset to quantify these aspects and calculated average scores. Aesthetics, a commonly employed metric in evaluating video dataset quality, was assessed automatically using the LAION Aesthetics Predictor~\cite{schuhmann2022laion}. The latter three metrics were chosen for their alignment with the criteria utilized in our qualitative analysis and user study (see \cref{sec:exp:human} and \cref{tab:human}). Consequently, we opted for human evaluation, with reviewers scoring each video on a four-point scale. Our dataset exhibits a substantial improvement across all four quality metrics compared to other datasets, thereby highlighting the superior quality of the videos.

\section{Method}\label{sec:method}
\subsection{Overview}
We provide a brief overview of latent diffusion models (LDMs) for text-to-video generation. LDMs extend traditional transformer models to handle the generative task and conduct the diffusion process in the latent space. First, a pre-trained autoencoder is utilized to compress the raw image or videos from pixel to latent space, and a text encoder takes the text input and creates text embeddings. A diffusion transformer (DiT) takes the visual input with noise and performs a denoising process during training. 
Specifically, as shown in \cref{fig:model}(a), the diffusion transformer consists of a stack of self-attention and cross-attention blocks, which capture the spatial and temporal dependencies within the video as long as text embeddings condition. During inference, the diffusion transformer starts from an instance sampled from random Gaussian noise and applies the diffusion process iteratively across multiple timesteps, refining the output at each step. 

To adapt the DiT model for long video generation, we modify the text embedding process and the cross-attention mechanism to effectively incorporate multiple progressive text conditions with temporal information. Specifically, we split the latent features into segments along temporal dimensions in the cross-attention, and study three different strategies to implement the interaction between text embedding and segmented latent features. The proposed \sca (SCA), especially the overlap variant, improves content richness and long-range coherence in generated long videos by a large margin. We will study this strategy and its different variants in the next subsection.

\subsection{Segmented Cross-Attention}

A standard paradigm of text-to-video generation relies on a single text prompt input, which is typically encoded into a fixed-sized embedding $c\in\mathbb{R}^{L\times D}$ via text encoder. Text embeddings that exceed this size are truncated. This limitation can lead to severe information loss in long video generation, considering the length of our progressive sub-captions. Moreover, a single long text embedding presents challenges in capturing intricate details, as latent representations within hidden states struggle to effectively capture the intricate details of a lengthy text embedding through cross-attention mechanisms.

Inspired by window attention\cite{liu2021swin}, which limits the scope of attention to local regions, we propose the Segmented Cross-Attention (SCA) method. This method splits the hidden states into temporal-local segments to better interact with the progressive sub-captions via cross-attention. 
For each group of progressive sub-captions, we separately encode $N$ sub-captions with text encoder, and thus obtain a group of text embeddings $\{c_i\}_{i=1}^{N}\in N\times\mathbb{R}^{L\times D}$. 
Given the hidden state $z$ with $T$ frames in the temporal dimension, we also evenly split $z$ into $N$ non-overlapped segments $\{z_i\}_{i=1}^{N}$ along the time dimension, aligning in quantity with the group of sub-captions. Thus, segment $z_i$ encapsulates the frame information ranging from $i\times T/N$ to $(i+1) \times T/N$. 
The core idea of our SCA is to restrict each segment $z_i$ to access only its corresponding text condition.

We study three strategies for segmented attention computing, as shown in \cref{fig:model}(b). For Isolate Segmented Cross-Attention (ISCA) shown in the first row of \cref{fig:model}(b), we treat it as the basic setting of our SCA, in which each segment $z_i$ only cross-attends to its corresponding text condition $c_i$ in cross-attention layers. 
Due to the lack of internal interactions between segments, ISCA tends to generate video with rich content but lacks long-range coherence. 

Another strategy is to enable long-range interactions between latent features and text embeddings, as shown in the second row of \cref{fig:model}(b). We refer to this variant as Sequential Segmented Cross-Attention (SSCA), which concatenates the latent segment $z_i$ sequentially with its latter segments $\{z_j\}_{j>i}^N$ and takes the average of all attention outputs in the final. Such a strategy greatly improves the long-range coherence. However, the content richness drops because the sequential interactive introduces more information to each segment and blends its content diversity.

Finally, we adopt a simple yet effective strategy that performs feature fusion on adjacent segments, which is Overlap Segmented Cross-Attention (OSCA), as shown in the third row of \cref{fig:model}(b). We relax the non-overlapping segment method above into the overlapping one by introducing $\delta < [T/N]$ overlapping frames for each segment $z_i$. Through overlapping, frames at the boundary of two adjacent segments will attend to multiple text conditions.
The cross-attention outputs in the overlapped regions are then averaged, promoting smoother transitions between segments.
OSCA allows each segment to cross-attend to its relevant text embeddings, while self-attention facilitates global information exchange across segments, ensuring overall consistency.
This interplay between local and global interactions helps \model effectively capture the storyline and ensures content diversity and scenario coherence in long-form video generation.

\subsection{Implementation}
Our work is built upon Allegro\cite{zhou2024allegro}, an open-source video diffusion model with 2.8B parameters. Allegro generates high-quality videos up to 88 frames and 720p resolution from simple text input. Text inputs are handled by T5~\cite{raffel2020exploring} text encoder. A video caption is decoupled into five progressive sub-captions and a hidden state is separated into five segments, corresponding to the notations $N$ above. For post-processing, we adopt EMA-VFI\cite{zhang2023extracting} as the frame interpolation model, to further normalize the video speed and extend video length. 
During inference, for a single prompt from user input, we leverage GPT-4o as the refiner to generate five progressive sub-captions.

The training of \model can be separated into two stages: Text-to-Video Pre-training, and Text-to-Video Fine-tuning.
The pre-training stage is built upon the Allegro model with 88 frames and $720\times1280$ resolution. The pre-training dataset contains 261k instances, and we sample frames from videos at 6 FPS during this stage. 
\model is trained for 1500 steps on 64 Nvidia H100 GPUs with a batch size of 256 and constant learning rate of 1e-4, processing a total of 384k videos. 
For fine-tuning, we pick the most content-diverse 47k instances from the pre-training dataset and fine-tune for another 500 steps with a batch size of 256 learning rate of 1e-4, processing a total of 128k videos.

\newlength{\myl}
\setlength{\myl}{4.8em}
\begin{table*}[htp]
  \centering
  \renewcommand{\arraystretch}{1.1}
  \renewcommand{\aboverulesep}{0pt}
  \renewcommand{\belowrulesep}{0pt}
  \setlength{\tabcolsep}{0pt}
  \resizebox{\linewidth}{!}{
  \begin{tabular}{p{7em}|P{\myl}P{\myl}P{\myl}P{\myl}P{\myl}P{\myl}|P{\myl}P{\myl}P{\myl}}
    \toprule
    \hspace{4pt} Dimensions & \multicolumn{6}{c|}{Specific Dimensions} & \multicolumn{3}{c}{Holistic Dimensions} \\
    \midrule
    \hspace{4pt} \multirow{2}{*}{Methods} & Dynamic  & Temporal & Human & Object & \multirow{2}{*}{Color}& Overall  &  Semantic & Quality & Overall \\
    & Degree  &  Style & Action & Class & & Consist. & Score & Score & Score \\
    \midrule
    \hspace{4pt} Gen-3 & 60.1 & \underline{24.7} & \bf 96.4 & \underline{87.8} & 80.9 & \underline{26.7} & \underline{75.2} & \bf 84.1 & \bf 82.3 \\
    \hspace{4pt} Allegro  & 55.0 & 24.4 & 91.4 & 87.5 & \underline{82.8} & 26.4 & 73.0 & \underline{83.1} &  \underline{81.1}\\
    \hspace{4pt} TALC &  \underline{98.6} & 18.0 & 89.0 & 45.3 & 57.3 & 19.5 & 44.4 & 62.5 & 58.9 \\
    \midrule
    \hspace{4pt} \model & \bf 100.0 & \bf 25.8 & \underline{93.0} & \bf 93.7 & \bf 98.1 & \bf 27.8 & \bf 78.5 & 80.6 & 80.2  \\
    \bottomrule
  \end{tabular}
  }
  \caption{Quantitative results of dimension performance on VBench. The \textbf{bold} means the best and the \underline{underline} means the second. We focus on the semantic dimension suite to demonstrate our \model is capable of generating content-rich videos with consistency.}
  \label{tab:vbench}
\end{table*}
\newlength{\myll}
\setlength{\myll}{3.6em}
\begin{table*}
  \centering
  \renewcommand{\arraystretch}{1.1}
  \renewcommand{\aboverulesep}{0pt}
  \renewcommand{\belowrulesep}{0pt}
  \setlength{\tabcolsep}{0pt}
  \resizebox{\linewidth}{!}{
  \begin{tabular}{p{7em}|P{\myll}P{\myll}P{\myll}|P{\myll}P{\myll}P{\myll}|P{\myll}P{\myll}P{\myll}|P{\myll}P{\myll}P{\myll}}
    \toprule
    \hspace{4pt} Dimensions &  \multicolumn{3}{c|}{Overall Score}  & \multicolumn{3}{c|}{Scenario Diversity}  & \multicolumn{3}{c|}{Scenario Coherence} & \multicolumn{3}{c}{Text-Video Adherence} \\
      \midrule
     \hspace{4pt}  Methods &  Win & Lose & Tie & Win & Lose & Tie & Win & Lose & Tie & Win & Lose & Tie  \\
      \midrule
     \hspace{4pt}  Gen-3 &  \bf 45.0 &  38.8 &  16.2 &  \bf 59.1 &  27.4 &  13.5 &  35.1 &  \bf 48.5 &  16.4 &  \bf 40.9 & 40.4 &  18.7  \\
    \hspace{4pt}   Allegro & \bf 54.9 & 27.0 &  18.1 & \bf 68.0 & 21.1 & 10.9 &  \bf  45.1 &  32.6 &  22.3 &  \bf  51.4 &  27.4 &  21.1  \\
   \hspace{4pt}    Merge Videos &  \bf 55.8 & 29.3 & 14.9 & \bf  45.5 &  44.8 &  9.7 &   \bf 71.5 &  18.8 &  9.7 &  \bf 50.3 &  24.2 &  25.5  \\
   \hspace{4pt}    TALC & \bf  91.8 &  3.1 & 5.1 & \bf  \bf 90.6 &  4.1 &  5.3 &  \bf  95.3 & 1.8 & 2.9 &  \bf 89.5 &  3.5 &  7.0  \\
    \bottomrule
  \end{tabular}
  }
  \caption{Qualitative results of win rate (\%) on user study. We ask users to evaluate two given videos based on three dimensions: Scenario Diversity, Scenario Coherence, and Text-Video Adherence. The Overall Score is calculated by considering all of the three dimensions.}
  \label{tab:human}
\end{table*}

\section{Experiment}
\label{sec:exp}
In this section, we demonstrate the long video generation capability of \model via both quantitative and qualitative evaluation in \cref{sec:exp:vbench} and \cref{sec:exp:human}, respectively. Moreover, we provide an ablation study in \cref{sec:exp:abla} on key components of our model, including training data, progressive sub-caption strategy, and \sca. We exhibit more results in \cref{app:qualitative} and \cref{app:control}, and discuss the limitations and failure cases in \cref{app:limitations}.

\subsection{Baseline Models}

To evaluate the effectiveness of our \model on content diversity and long-range coherence, we compare it with the state-of-the-art text-to-video models. 
We select the best open-source model, Allegro~\cite{zhou2024allegro}, and commercial system, Runaway Gen-3~\cite{gen3}, as our baseline models. 
We also compare with the recent MT2V method, TALC~\cite{bansal2024talc}. 
To highlight the importance of scenario coherence, we add a naive approach of ``Merge Videos" in qualitative evaluation, by utilizing multiple texts to generate multiple short clips.

\subsection{Quantitative Evaluation}\label{sec:exp:vbench}
\noindent\textbf{Setup} We use VBench for our automatic quantitative evaluation. 
VBench offers 946 official prompts to validate different aspects of generated videos and is a widely adopted benchmark in video generation methods. 
To align with the evaluation of other models, we assess the original videos with 88 frames before interpolation for our \model. 

We directly report the results for models that exist in the VBench Leaderboard. 
We report several specific dimensions and the holistic dimensions in VBench, as shown in \cref{tab:vbench}, demonstrating the exceptional capability of generating long videos with rich content and long-range coherence while adhering to the input text.

\model outperforms all state-of-the-art video generation models on Semantic Score. 
Specifically, \model notably surpasses Allegro with $+5.5\%$, the commercial Gen-3 with $+3.3\%$, and the previous MT2V method TALC with $+34.1\%$.
We attribute the performance improvement to the progressive sub-captions generation strategy, which decouples the text input and improves the text information.
Besides, we achieve a \textit{full mark} in Dynamic Degree metrics, reflecting the superior ability to capture dynamics and preserve camera control. 
Compared with TALC, which achieves a relatively high score of $98.6\%$ in Dynamic Degree, we achieve significantly better performance on all metrics, highlighting the long-range coherence aided by the meticulous data curation and \sca.
For the degradation in quality scores, we hypothesize that it arises from the increased difficulty in maintaining consistency when the dynamics are complex and varied. 

We explore more details of the hypotheses made within this sub-section in the Appendix, including `\textit{progressive sub-captions contribute to higher semantics}' in \cref{app:prog}, and `\textit{complex dynamics lead to degradation in quality}' in \cref{app:dyn}.

\begin{figure*}
\begin{center}
\centerline{\includegraphics[width=0.98\linewidth]{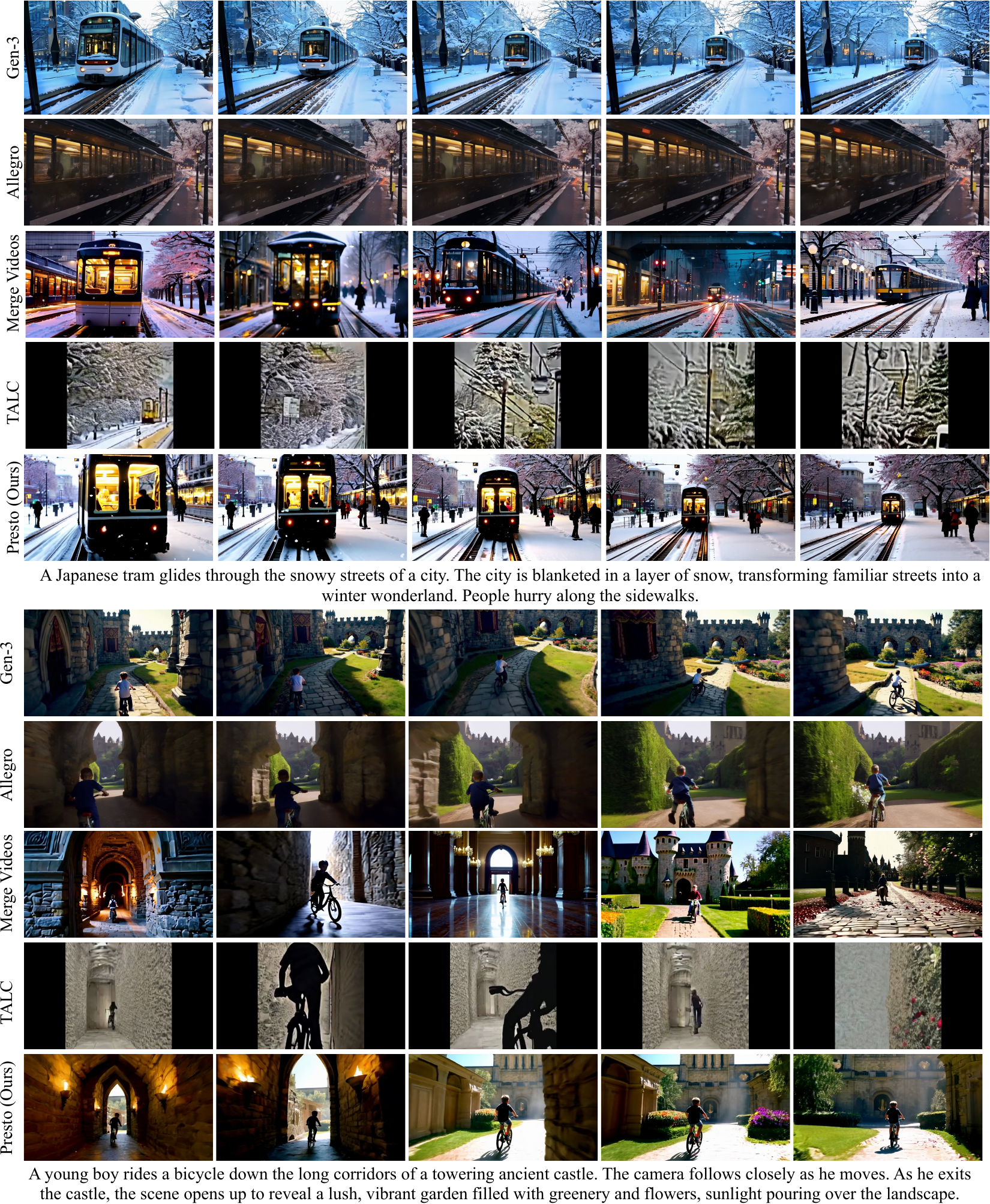}}
\vspace{-5pt}
\caption{Qualitative comparison with the baselines in our user study. Our \model can capture intricate text details and generate long videos with long-range coherence and rich content. For the first case, ours is the only method that captures the text details of \textit{``People hurry along the sidewalk"}, while other methods fail to generate walking people. For the second case, our generated videos are of the largest camera motion and the best scenario coherence.}
\label{fig:compare}
\end{center}
\end{figure*}

\subsection{Qualitative Evaluation}\label{sec:exp:human}

\noindent\textbf{Setup} Evaluating the quality of generated videos is a highly subjective task, as automatic benchmarks often dis-align with human judgment. 
A user study is a prevalent paradigm for qualitative assessment in previous work \cite{yang2024cogvideox,zhou2024allegro,zheng2024contphy}.
In this study, we collected 62 diverse text prompts covering a wide range of aspects, including humans, animals, landscapes, and more.
Human annotators are tasked with blindly comparing pairs of videos and making a preference judgment between two cases.
To assess the enhanced content diversity achieved by our model, we evaluate three key dimensions: 1) Scenario Diversity, which video is more diverse considering the changing scenario; 2) Scenario Coherence, which video is more coherent on maintaining objects and background in the changing scenario; and 3) Text-Video Adherence, which video is closer to the user prompt.
We allow the tie situation when the differences are indistinguishable.
We recruited 12 annotators, and each instance was reviewed by three individuals, resulting in 2,232 ratings. 
We report the win rate (\%) with each method and calculate an overall score of three dimensions in \cref{tab:human}.

Our model surpasses all baselines on the Overall Score, indicating that our \model can generate videos with improved scenario diversity and coherence with text-following capabilities, even better than the commercial model Gen-3.
Specifically, \model excels Gen-3 in scenario diversity, slightly outperforms in text-video adherence, and closely matches in scenario coherence. For the SOTA open-source model, Allegro, \model wins in all dimensions, especially on the scenario diversity. For TALC, \model significantly outperforms in all metrics, further demonstrating the importance of curated data. 
We also consider the naive approach of utilizing multiple texts by ``Merging Videos", to serve as a reference for our metrics.
Our method reaches similar results as Merging Videos on scenario diversity but significantly outperforms it in scenario coherence. 

We further exhibit the real cases in our user study, as shown in \cref{fig:compare}. For the first case, our \model is the only one that captures intricate text details of ``\textit{People hurry along the sidewalk.}" while all other methods fail to generate walking people. For the second case, our \model achieves the largest scenario motion while with the best long-range coherence. Merge video may generate highly diverse content, but it fails to maintain consistency as five shots are different and unrelated. 

\subsection{Ablation Study}\label{sec:exp:abla}

\noindent We now ablate on the key proposed components, including both model design and dataset curation. We use VBench to evaluate generated videos on three dimensions: 1) Overall Score; 2) Dynamic Degree; and 3) Overall Consistency. 
As video generation methods consume huge computational resources in model training, we standardized our video generation model training to 360p resolution with 40 frames in the ablation study. \cref{tab:abla} presents the ablation results on the automatic evaluation benchmark.

\noindent\textbf{\sca (SCA) Strategy.} 
We ablate the three strategies of SCA, including Overlap \sca (OSCA), Sequential \sca (SSCA), and Isolated \sca (ISCA), as we mentioned in \cref{sec:method} and \cref{fig:model}(b).
All three strategies achieve 100\% in the Dynamic Degree score, indicating that the general concept of SCA plays a significant role in capturing dynamics.
OSCA achieves an excellent result of 74.7\% Overall Score and 25.29\% Overall Consistency, which is the best version among all strategies. This design not only maintains content richness but also facilitates transitions between segments by introducing proximal overlapping.
SSCA is slightly behind OSCA, and we attribute this degradation to the information from previous frames disrupting the interaction of later segmented latent features. 
ISCA performs poorly in experiments, as the isolated design prohibits information exchange in adjacent segments.
This ablation further underscores the significance of the overlapping technique in OSCA, facilitating smoother transitions between segments.

\newlength{\aw}
\settowidth{\aw}{$\downarrow$}
\newlength{\mw}
\settowidth{\mw}{$-$}
\newlength{\hackw}
\setlength{\hackw}{5pt}

\begin{table}
  \centering
  \renewcommand{\arraystretch}{1.1}
  \renewcommand{\aboverulesep}{0pt}
  \renewcommand{\belowrulesep}{0pt}
  \setlength{\tabcolsep}{0pt}
  \resizebox{\columnwidth}{!}{
  \begin{tabular}{p{11em}|P{6em}|P{6em}|P{6em}}
    \toprule
     \hspace{4pt} Method & Overall &  Dynamic & Consistency  \\
     \midrule
   \hspace{4pt} O(verlap) SCA & 74.7 & 100.0 & 25.29  \\
    \rowcolor{lightgray} \multicolumn{4}{c}{\textit{Segmented Cross-Attention (SCA) Strategy}} \\
   \hspace{4pt}  S(equential) SCA & \makebox[\aw]{} 73.7 $\downarrow$ & 100.0 & \makebox[\aw]{} 25.06 $\downarrow$ \\
   \hspace{4pt}  I(solated) SCA & \makebox[\aw]{} 73.1 $\downarrow$ & 100.0 & \makebox[\aw]{} 24.88 $\downarrow$ \\
    \rowcolor{lightgray} \multicolumn{4}{c}{\textit{\dataset Dataset Curation}} \\
    \hspace{4pt} w/o Meticulous Filtering & \makebox[\aw]{} 72.0 $\downarrow$   & \makebox[\aw]{} 97.2 $\downarrow$  & \makebox[\aw]{} 24.06 $\downarrow$ \\
    \hspace{4pt} Single Long Condition & \makebox[\aw]{} 71.8 $\downarrow$  & 100.0 & \makebox[\aw]{} 24.06 $\downarrow$  \\
    \bottomrule
  \end{tabular}
  }
  \caption{Ablation results of model design and dataset curation.}
  \label{tab:abla}
\end{table}

\noindent\textbf{w/o Meticulous Filtering.} 
We study the effect of our meticulously curated dataset \dataset, with the same strategy of OSCA.
We randomly select the same amount of pre-training data from the video datasets without the filtering of content diversity in \cref{sec:data:videos} and captions in \cref{sec:data:caption}. 
Note that we still adopt basic filtering for videos, including duration, speed, resolution, and low-level metrics, to ensure basic visual quality and reduce the effect of metrics contributing to rich content.
The model with worse data achieves 72.0\% on the overall score, largely lagging behind 2.7\% in the same strategy with well-curated data. 
Moreover, the performance on Dynamic Degree and Overall Consistency will also drop by 2.8\% and 1.23\%, respectively, highlighting the effectiveness of our data curation.

\noindent\textbf{Single Long Condition.} 
To diminish the effect of the increased text length in our sub-captions and demonstrate the effectiveness of separate modeling in our \sca, we test the method of naive concatenation on text condition. Specifically, this approach directly concatenates $N$ sub-captions along the sequence length dimension, forming a single long text condition. This long condition will attend to the whole hidden states, the same as standard text-to-video generation approaches. 
Results show that this naive concatenation method yields the lowest Overall Score with a 2.9\% drop, indicating that our \sca effectively enhances the overall video quality.

\section{Conclusion}
\label{sec:con}

We introduced \model, a simple yet effective method for generating long-range coherent, content-rich, long videos.
Our \model achieves 78.5\% and 100\% on the VBench Semantic Score and Dynamic Degree, surpassing the existing SOTA video generation approaches.
\model utilizes the Segmented Cross Attention mechanism to integrate multiple texts concurrently, which can be seamlessly adopted in the existing diffusion model with DiT architecture. 
We also curate a high-quality video-texts dataset \dataset from public sources. 
We demonstrate that high quality video-text pairs are crucial for long video generation, and our curated \dataset is a good candidate for future research.
We leave more exploration about the attention mechanism and model structure (\eg, auto-regressive generation) for long video generation as our future work.

\section*{Acknowledgement}

The authors appreciate Wei Chen, Huiguo He, Mindy Lin, Zeyu Liu, Yuhang Zhang, Yuanzhi Zhu for their valuable input and suggestions.

{
    \small
    \bibliographystyle{ieeenat_fullname}
    \bibliography{main}
}

\begin{appendix}

\clearpage
\maketitleappendix

\section{Details of \dataset Dataset}\label{app:details}

\begin{figure*}
\begin{center}
\centerline{\includegraphics[width=0.98\linewidth]{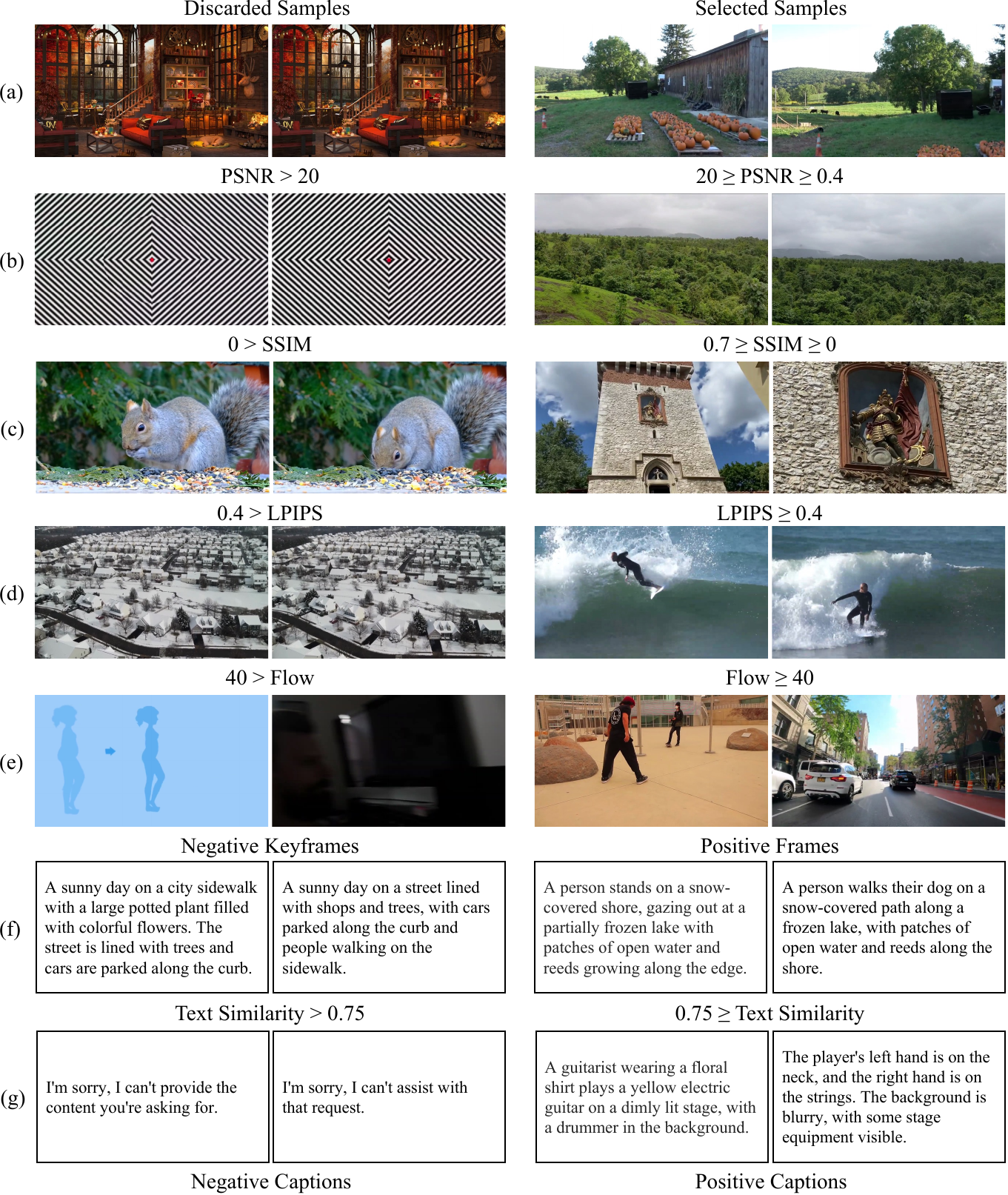}}
\caption{The discarded and selected data samples of different filtering steps in \dataset. We discard cases with similar keyframes and poor content diversity and filter out similar and negative captions. The selected cases have rich video content, coherent scenario motion, and progressive captions. We visualize the samples in the \dataset Pre-training set and apply more rigorous filtering to develop the \dataset Fine-tuning set.
}
\label{fig:data}
\end{center}
\end{figure*}

\begin{figure*}
\begin{center}
\centerline{\includegraphics[width=0.98\linewidth]{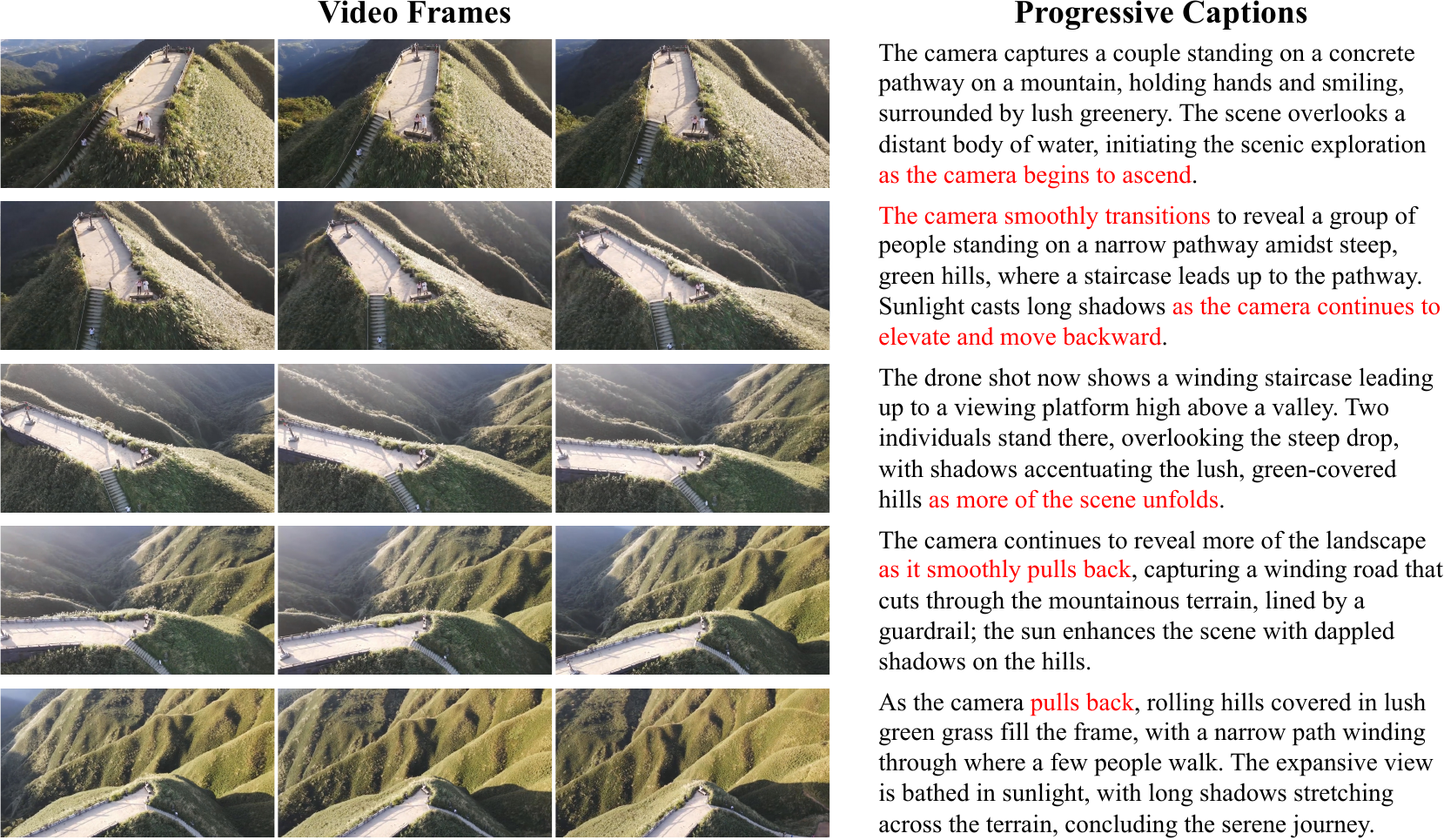}}
\caption{The progressive sub-captions and coherent video frames of our \dataset dataset. Our captions are more detailed in camera motion, as highlighted in the red text.}
\label{fig:case}
\end{center}
\end{figure*}

In this section, we show more details of our filtering steps, contributing to the \dataset dataset with rich content and long-range coherence. Thresholds for each step are displayed in \cref{tab:data}. We visualize the discarded samples and selected samples of each filtering step in \cref{fig:data}. Moreover, we exhibit a real case with coherent video frames and progressive captions in our \dataset in \cref{fig:case}.

\noindent\textbf{Pixel-wise Filtering.} We use the Peak Signal-to-Noise Ratio (PSNR) to ensure the sampled keyframes are pixel-wisely diverse and coherent. We filter out the cases with high PSNR values, indicating the keyframes are not diverse enough, as visualized in \cref{fig:data}(a).

\noindent\textbf{Structure-wise Filtering.} We employ the Structural Similarity Index Measure (SSIM) to measure the structural-wise similarity of the keyframe diversity. We filter out similar cases with higher SSIM values, and the cases with SSIM values lower than 0, which indicates that the image structures are inverted~\cite{wang2005structural}, as visualized in \cref{fig:data}(b).

\noindent\textbf{Semantics-wise Filtering.} We adopt the Perceptual Similarity (LPIPS) to evaluate the semantic diversity and coherence of sampled keyframes. We visualize a discarded case and selected case in \cref{fig:data}(c).

\noindent\textbf{Motion-wise Filtering.} We utilize Unimotion to calculate the optical flow values of each video clip per second. Videos with higher flow values are both coherent and dynamic across scenarios, as visualized in \cref{fig:data}(d).

\noindent\textbf{Text-wise Filtering.} We utilize Aria~\cite{li2024ariaopenmultimodalnative} as our captioning model, and utilize MPNet~\cite{song2020mpnet} from SentenceTransformers~\cite{reimers-2020-multilingual-sentence-bert, reimers-2019-sentence-bert} to compute the cosine similarity~\cite{singhal2001modern} of each text pair. We filter out the cases with higher text similarity, as displayed in \cref{fig:data}(f), to enhance the diversity in text captions. We further utilize GPT-4o~\cite{achiam2023gpt} as the LLM for refining the sub-captions. Prompt templates for these two steps are displayed in \cref{lst:aria} and \cref{lst:gpt}.

\noindent\textbf{Negative Cases.} We show the negative cases of keyframes and captions in \cref{fig:data}(e) and \cref{fig:data}(g) respectively. 
Blurry or unrelated keyframes are discarded, by analyzing the compressed image file size.
Negative captions with sensitive information or when LLMs refuse to respond will be filtered out to improve the quality of captions.

\begin{table}
  \centering
  \renewcommand{\arraystretch}{1.1}
  \renewcommand{\aboverulesep}{0pt}
  \renewcommand{\belowrulesep}{0pt}
  \setlength{\tabcolsep}{0pt}
  \resizebox{\columnwidth}{!}{
  \begin{tabular}{p{10em}|P{7em}P{7em}}
    \toprule
    \hspace{4pt} Filtering & Pre-training & Fine-tuning \\
    \midrule
    \rowcolor{lightgray} \multicolumn{3}{c}{\textit{Content-Diverse Video Clips}} \\
    \hspace{4pt} Width & $\geq1280$ & $\geq1280$ \\
    \hspace{4pt} Height & $\geq720$ & $\geq720$ \\
    \hspace{4pt} FPS & $[24, 60]$ & $[24, 60]$ \\
    \hspace{4pt} Duration & $\geq 15$ & $\geq 15$ \\
    \hspace{4pt} Grayscale & $[20, 180]$ & $[20, 180]$ \\
    \hspace{4pt} LAION Aesthetics & $\geq 4.8$ & $\geq 5.0$ \\
    \hspace{4pt} Tolerance Artifacts & $\leq 5\%$ & $\leq 5\%$ \\
    \hspace{4pt} Unimatch Flow & $\geq 40$ & $\geq 50$ \\
    \rowcolor{lightgray} \multicolumn{3}{c}{\textit{Coherent Video Captions}} \\
    \hspace{4pt} PSNR & $[4, 20]$ & $[4, 20]$ \\
    \hspace{4pt} SSIM & $[0, 0.7]$ & $[0, 0.7]$ \\
    \hspace{4pt} LPIPS & $\geq 0.4$ & $[0.5, 0.8]$ \\
    \hspace{4pt} Text Similarity & $\leq 0.75$ & $[0, 0.75]$ \\
    \bottomrule
  \end{tabular}
  }
  \caption{Data filtering thresholds across various stages. All thresholds are manually determined by the specific characteristics of the dataset.}
  \label{tab:data}
\end{table}

\begin{table}
  \centering
    \renewcommand{\arraystretch}{1.1}
  \renewcommand{\aboverulesep}{0pt}
  \renewcommand{\belowrulesep}{0pt}
      \resizebox{0.99\linewidth}{!}{
  \begin{tabular}{l|cccc}
    \toprule
     Sub-captions & Similarity $\uparrow$ & ROUGE-L $\uparrow$ & BLEU-4 $\uparrow$ \\
      \midrule
    Vanilla & 0.6408 & 0.1968 & 0.0376 \\
    Progressive & \bf 0.7778 & \bf 0.2306 & \bf 0.0578 \\
    \bottomrule
  \end{tabular}
  }
  \caption{Text similarity of training captions and inference captions, compared between vanilla style and progressive style.}
  \label{tab:textsim}
\end{table}

\section{Details of Progressive Sub-captions}\label{app:prog}
Progressive sub-captions have been demonstrated to improve semantic scores in diffusion model training~\cite{villegas2022phenaki}. Intuitively, the progressive style enhances caption coherence, mitigating redundant information and phrasing. This section offers a unique perspective to further substantiate this argument: the LLM-refined progressive style outperforms the non-refined vanilla style for training sub-captions. We adopt a text-centric approach, evaluating this hypothesis by computing the text similarity between training and inference captions (note that inference captions remain constant). This experimental design comes from the intuitive notion that closer distribution between inference and training data will yield better results. We employ Cosine Similarity~\cite{reimers-2020-multilingual-sentence-bert, reimers-2019-sentence-bert}, Rouge-L~\cite{lin2004automatic,lin2004rouge}, and BLEU-4~\cite{papineni2002bleu} metrics to assess the text similarity. As evidenced in \cref{tab:textsim}, progressive-style captions exhibit improved text similarity compared to vanilla-style captions across all metrics, indicating a better semantic score in the generated videos. This observation indirectly validates our hypothesis.

We acknowledge that the most direct validation would involve training diffusion models under identical settings with both captioning styles and subsequently comparing the quality of the generated videos. However, given the substantial computational resources required for diffusion model training, we reserve this comprehensive evaluation for future work.

\section{Analysis of Videos with Complex Dynamics}\label{app:dyn}

This section mainly analyzes the reasons for the quality degradation in videos with complex dynamics. We refer to the issue of high-dynamism training videos suffering from quality degradation

This section primarily investigates the underlying causes of quality degradation observed in videos exhibiting complex dynamics. We term the phenomenon of highly dynamic training videos experiencing quality degradation as `dynamism loss'. Several factors contribute to this effect: 1) Individual frames within dynamically complex videos are inherently more susceptible to motion blur; and 2) Video format compression, specifically H.264 encoding employed in our experiments, induces greater quality loss in videos with higher dynamic range. This `dynamism loss' happens twice when generating a content-rich video in our experiments, occurring both during the filtering and transcoding of training video data, and subsequently during the saving and encoding of generated videos. This two-fold occurrence accounts that it's harder to maintain the same level of video quality compared with dynamic videos and normal videos, thus explaining the observed quality decline in our generated videos. 

\section{More Qualitative Comparisons}\label{app:qualitative}

\begin{figure*}
\begin{center}
\centerline{\includegraphics[width=0.98\linewidth]{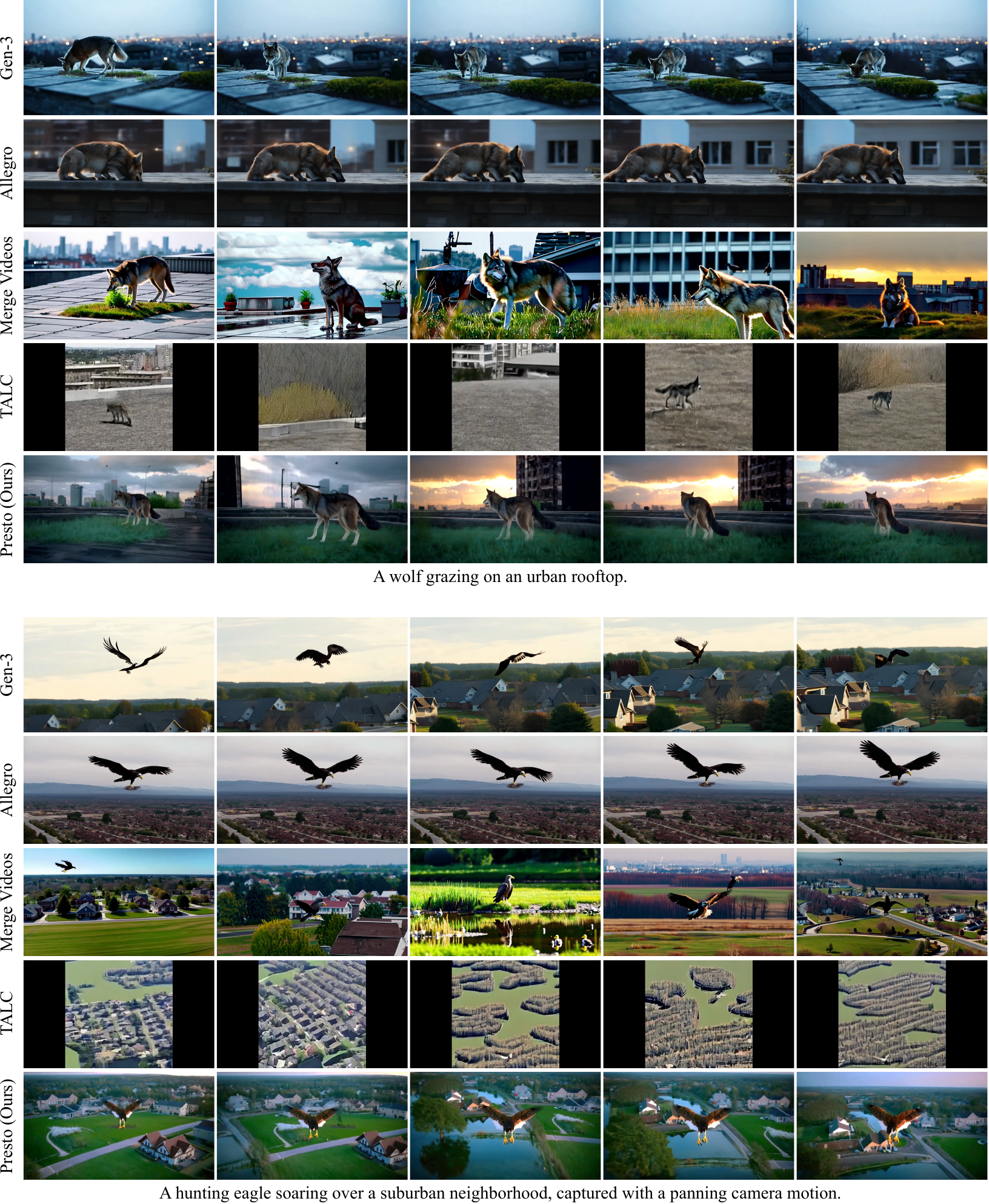}}
\caption{Qualitative comparison with the baselines in our user study. }
\label{fig:comp2}
\end{center}
\end{figure*}

\begin{figure*}
\begin{center}
\centerline{\includegraphics[width=0.98\linewidth]{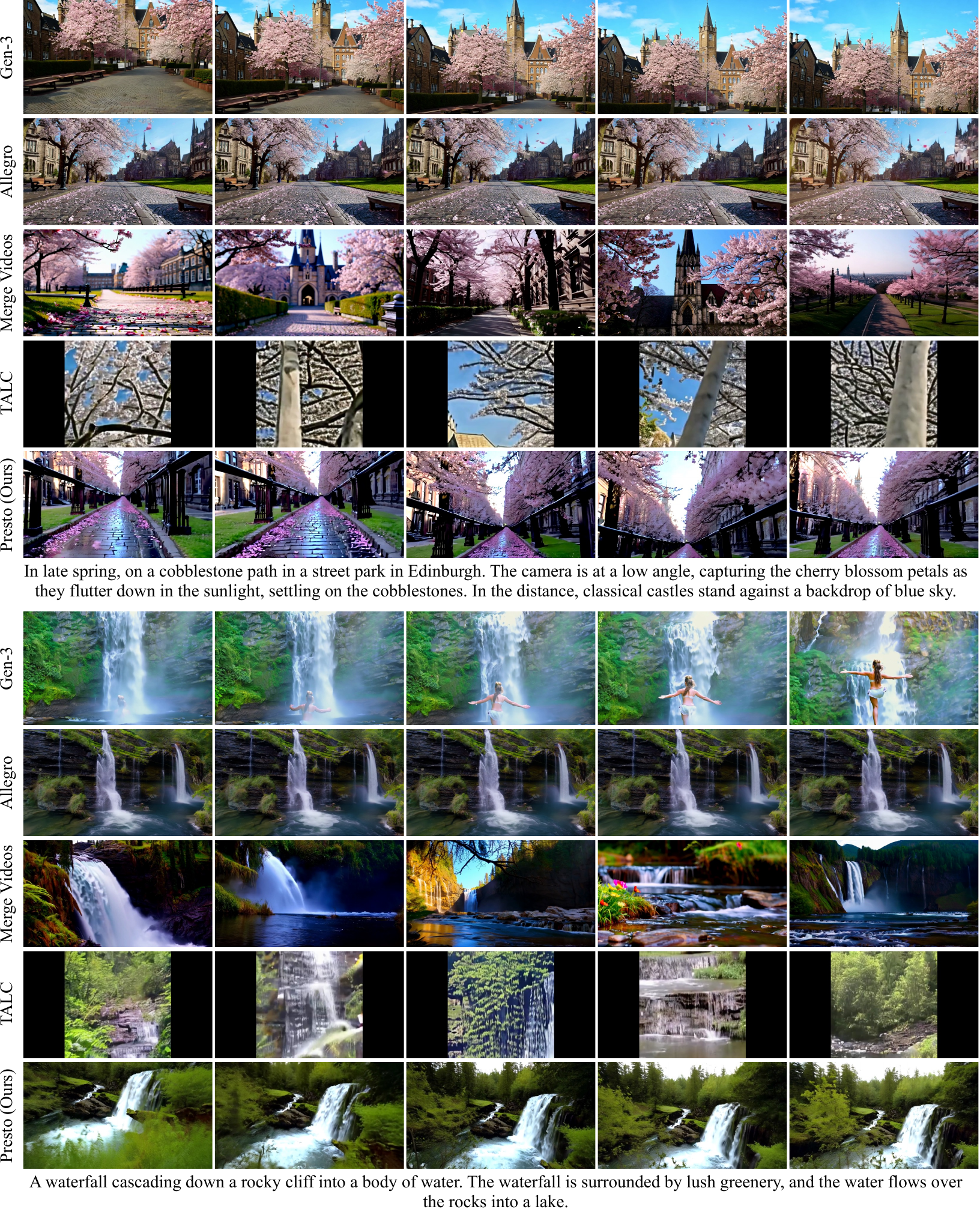}}
\caption{Qualitative comparison with the baselines in our user study.}
\label{fig:comp3}
\end{center}
\end{figure*}

We show more qualitative results compared with different baselines in \cref{fig:comp2} and \cref{fig:comp3}. Our generated videos have the largest scenario motion and maintain long-range coherence. 

\section{Style Control and Camera Control}\label{app:control}

\begin{figure*}
\begin{center}
\centerline{\includegraphics[width=0.98\linewidth]{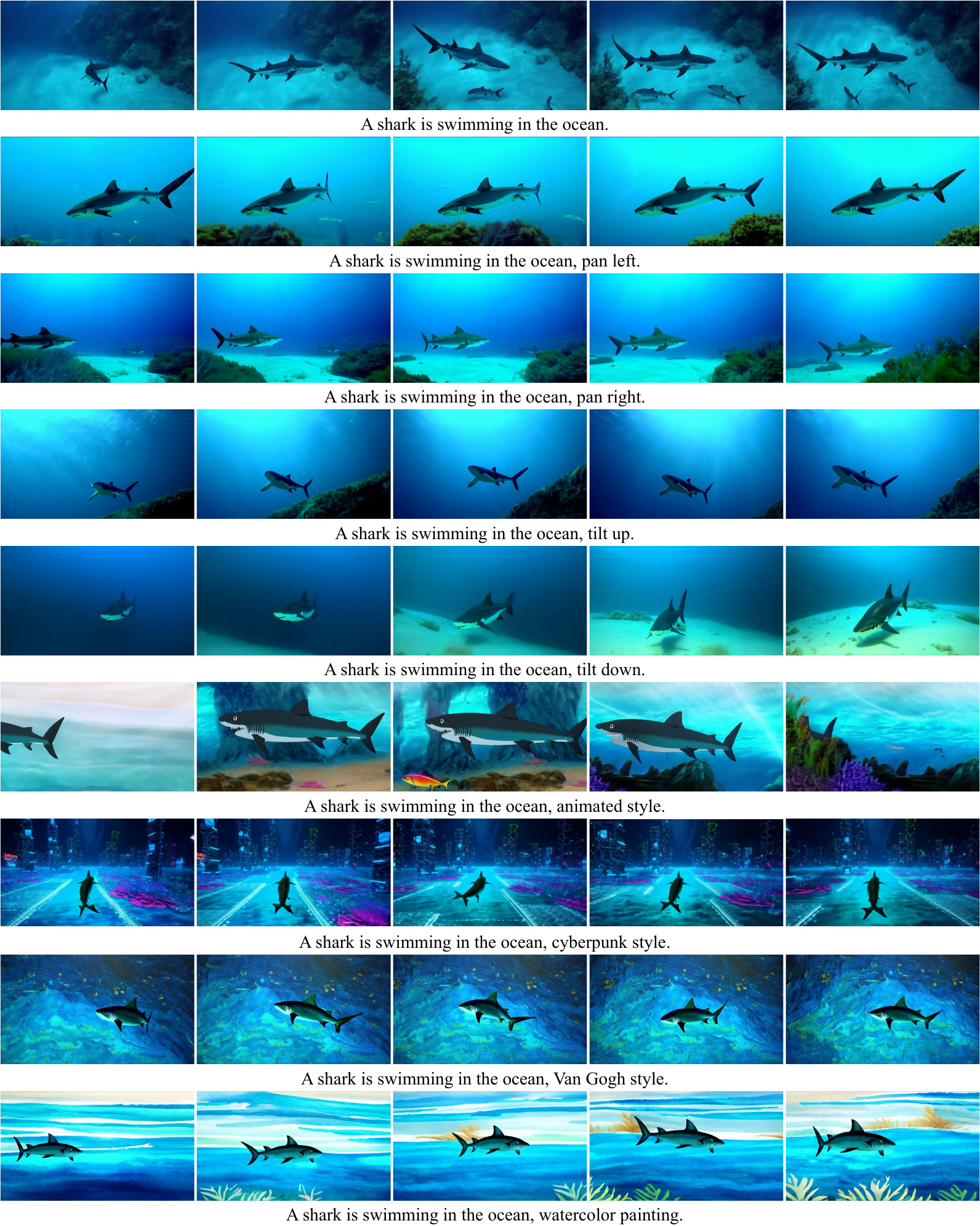}}
\caption{More results of VBench's prompts centering around the same theme. \model can generate videos with accurate camera control and style control.}
\label{fig:more}
\end{center}
\end{figure*}

To exhibit the superior capability of style control and camera control of our proposed \model, we select a series of prompts from the VBench, all centered around the same theme, `A shark is swimming in the ocean', but with variations in camera poses and styles. 
As shown in \cref{fig:more}, the results demonstrate that our model accurately adheres to the style and camera specifications provided in user input text.

\section{Limitations}\label{app:limitations}

\begin{figure*}
\begin{center}
\centerline{\includegraphics[width=0.8\linewidth]{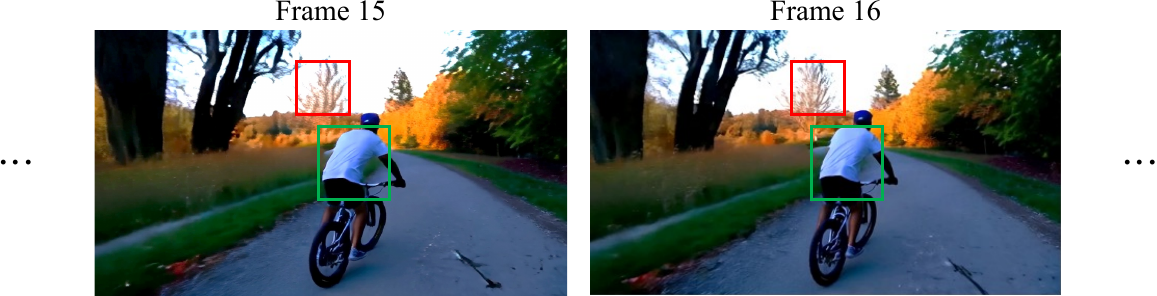}}
\caption{Our \model can generate long videos with high scenario motion, and prioritize scenario smoothness. However, in the case of extreme scenario motion, the main object will retain details and sharpness (as shown in the green box), while the moving background makes it easier to display artifacts such as blurring or ghosting (as shown in the red box). }
\label{fig:failure}
\end{center}
\end{figure*}

Although our proposed \model can generate long videos with long-range coherence and rich content, certain limitations remain. First, the generated videos sometimes slightly degrade visual fidelity compared to the base model. 
We attribute this to the exclusive use of publicly accessible videos for training, which, while diverse and coherent, still do not match the higher quality of the private datasets leveraged by the base model.
Second, in cases involving extreme scenario motion, some regions may display artifacts such as blurring or ghosting, as visualized in \cref{fig:failure}. These artifacts are likely a consequence of our model prioritizing scenario consistency and smoothness, which occasionally compromises spatial sharpness in high-motion backgrounds.
Last, our model is not suitable for generating still frames.

\begin{figure*}[t]
\begin{lstlisting}[language={Tex},xleftmargin=.1\textwidth, xrightmargin=.1\textwidth,firstnumber=1, caption=Prompt template for video and image captioning., label={lst:aria}]
% Prompt Template for Image Caption
<IMAGE>
Describe the image in as much detail as possible. Incorporate the alt text if it provides information related to the visual scene.
alt text: <ALT_TEXT>

% Prompt Template for Video Caption
Write a concise, continuous prompt describing the video for generation, including objective facts, main subjects, their movements and positions, interactions, human actions, data sources, lighting, environment, camera angles, movements, background, atmosphere, photography style, fashion, and temporal information. Use professional or simple language for camera angles and movements.
<VIDEO>
\end{lstlisting}
\end{figure*}
\begin{figure*}[t]
\begin{lstlisting}[language={Tex},xleftmargin=.1\textwidth, xrightmargin=.1\textwidth,firstnumber=1, caption=Prompt Template for GPT-4o Refinement., label={lst:gpt}]
% Prompt Template for Sub-captions Refinement in LongTake-HD Dataset
System Prompt: 
You are a helpful video director. Refine the five scene descriptions to become more coherent based on the provided five frame desciptions and the video description.

User Prompt:
I will show you five scene descriptions in progressive frame level, as well as the video description. The refinement should follow these rules: 
1. Refinement should be based on the corresponding frame description, and can add information based on the video description. Do NOT imagine or add other new information. Do NOT change the order of each description.
2. There needs to be connections between the five scenes. Analyze the scenario transitions (such as camera movement, background changes, and object movement), and add them to each description. The camera movement should be smooth. 
3. The five scenes must form a continuous story, which means repeated object descriptions and details may be omitted. You need to accurately, objectively, and succinctly describe everything. The scene descriptions need to be concise. Do NOT add too many details unrelated to the video content description. 
4. Frame descriptions are independent, so there may be duplication. You need to analyze the possible states of different frames based on the video description. Do NOT incorporate later details into the previous frame's description.
The whole video description: <VIDEO_CAPTION>
Five descriptions at different frames: <FRAME_CAPTIONS>

% Prompt Template for Sub-captions Generation in Inference Stage
System Prompt: 
You are a helpful video director.

User Prompt:
Based on the video content description, you need to write five coherent scene descriptions to create a silent video. These five descriptions are independent, but there needs to be a connection between the five scenes. The five scene descriptions should include detailed scenario transitions (such as camera movement, background changes, and object movement). The camera movement should be smooth. Avoid drastic angle changes and transitions, such as shifting from a frontal view directly to a side view. You can add details and objects, but the five scenes must form a continuous story, which means repeated object descriptions and details may be omitted. Five scene descriptions should NOT differ too much. Ensure similarity to enable smooth transitions between scenes. If the description is brief, you can add details, but stay conservative, and only create simple, easily generated scenes. It's also acceptable for multiple scenes to share a higher degree of similarity. You need to accurately, objectively, and succinctly describe everything. The scene descriptions need to be concise. Do NOT add details unrelated to the video content description. Do NOT speculate. Do NOT add scene titles, directly return five scene descriptions. 
The video content description: <VIDEO_DESCRIPTION>
\end{lstlisting}
\end{figure*}

\end{appendix}

\end{document}